\documentclass[letterpaper, 10 pt, conference]{ieeeconf}  % Comment this line out if you need a4paper

\IEEEoverridecommandlockouts                              % This command is only needed if 
\usepackage{graphicx} % for pdf, bitmapped graphics files
\usepackage{caption}
\usepackage{subcaption}
\usepackage{amsmath}
\usepackage{amsfonts}
\usepackage[cal=rsfso,calscaled=.96]{mathalfa}
\usepackage{mathrsfs}  
\usepackage{algpseudocode}
\usepackage[ruled,vlined]{algorithm2e}

\title{\LARGE \bf
Self-learning and adaptation in a sensorimotor framework
}

\author{Ali Ghadirzadeh,  Judith B\"utepage, Danica Kragic and M{\r a}rten Bj\"orkman% <-this % stops a space
\thanks{Authors are with the Computer Vision and Active Perception Lab (CVAP), CSC, KTH Royal Institute of Technology, Stockholm, Sweden.
        {\tt\small algh|butepage|dani|celle@kth.se}}%
}

\begin{document}

\maketitle
\thispagestyle{empty}
\pagestyle{empty}

%%%%%%%%%%%%%%%%%%%%%%%%%%%%%%%%%%%%%%%%%%%%%%%%%%%%%%%%%%%%%%%%%%%%%%%%%%%%%%%%
% 		TEXTS
%%%%%%%%%%%%%%%%%%%%%%%%%%%%%%%%%%%%%%%%%%%%%%%%%%%%%%%%%%%%%%%%%%%%%%%%%%%%%%%%
\begin{abstract}

We present a general framework to autonomously achieve a task, where autonomy is acquired by learning sensorimotor patterns of a robot, while it is interacting with its environment. To accomplish the task, using the learned sensorimotor contingencies, our approach predicts a sequence of actions that will lead to the desirable observations. 

Gaussian processes (GP) with automatic relevance determination is used to learn the sensorimotor mapping. In this way, relevant sensory and motor components can be systematically found in high-dimensional sensory and motor spaces. We propose an incremental GP learning strategy, which discerns between situations, when an update or an adaptation must be implemented. RRT* is exploited to enable long-term planning and generating a sequence of states that lead to a given goal; while a gradient-based search finds the optimum action to steer to a neighbouring state in a single time step.  

Our experimental results prove the successfulness of the proposed framework to learn a joint space controller with high data dimensions (10$\times$15). It demonstrates short training phase (less than 12 seconds), real-time performance and rapid adaptations capabilities. 

\end{abstract}

\section{INTRODUCTION}
% ----  intro 
Sensorimotor learning is a vital ability resulting in skilled performance in biological systems.
However, the amount of uncertainty presented in both sensory and motor channels
impedes the learning of even basic actions significantly. 
Additionally, an embodied agent has to control and optimize trajectories in high-dimensional 
sensory and motor spaces in a changing and dynamic environment, that further complicates learning. 
In humans, the interplay of sensory and motor signals is the substantial basis to allow movement 
generation under these complicated conditions \cite{franklin11}.

Biological studies \cite{wolpert11} suggest three main categories of control mechanisms - reactive, 
predictive and biomechanical.    
As latency in the processing of sensory data is inherent in these systems,
predictive control plays a significant role in skilled action generation.
It is based on the learning of a mapping between motor commands and sensory observations.  
This mapping, known as a forward model, predicts the sensory outcomes of a given action. 

\begin{figure}[t!]
      \centering
      \includegraphics[width=0.44\textwidth]{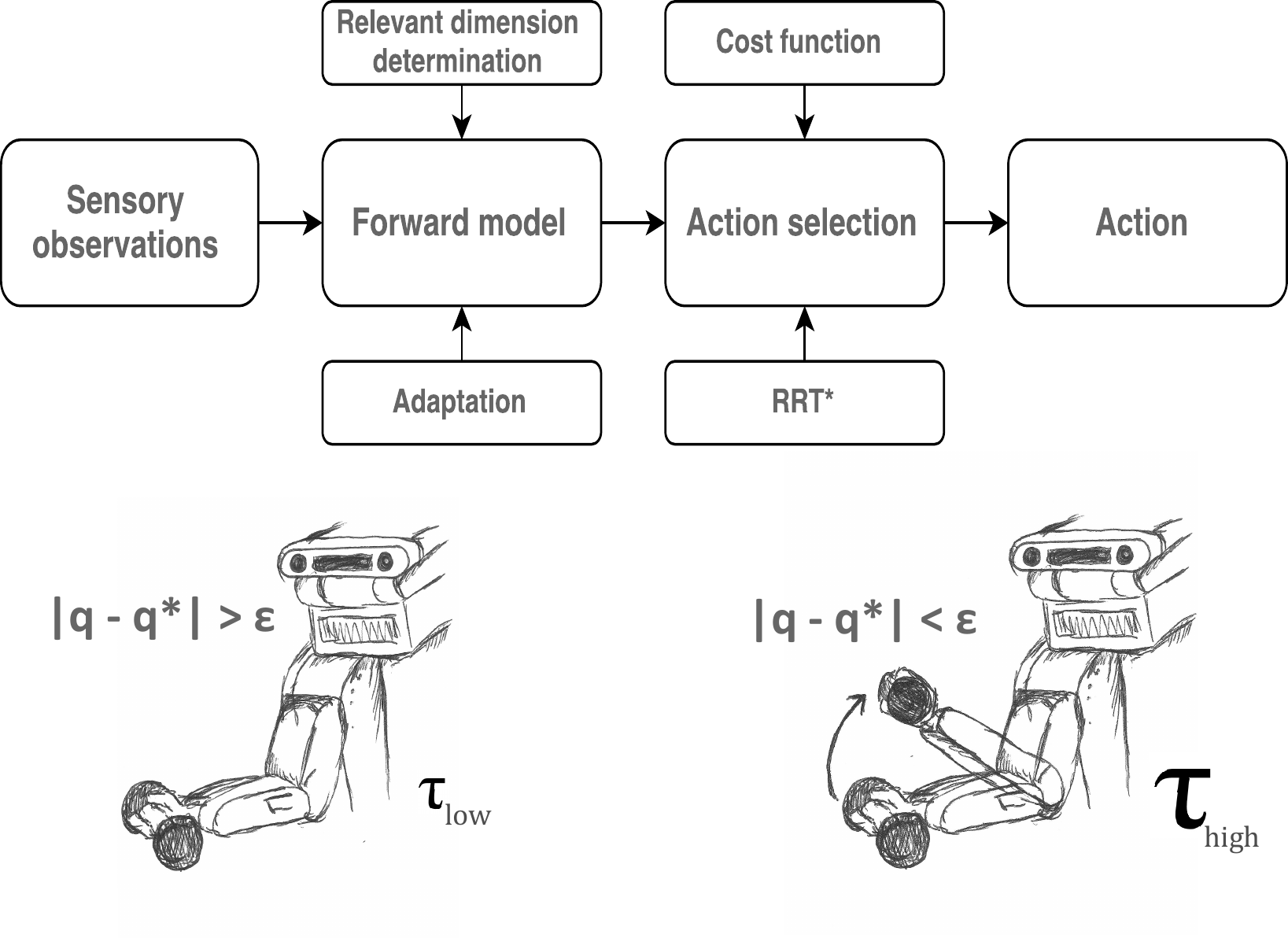}
      \caption{Top: The sensorimotor framework - 
        The forward model provides the basis for relevant dimension selection and adaptation. 
        Gradient-based optimization and the RRT* algorithm enable the selection of proper actions
		by optimizing a non-convex cost functions. 
      	Bottom: The perceived mismatch between the predicted ($q^*$) and actual ($q$) joint 
		position is larger than a threshold ($\epsilon$). 
		This suggests that the system should adapt accordingly to produce a larger effort.}		
      \label{fig:intro}
\end{figure}

Additionally, the forward model is essential in
error-based learning and mismatch detection - two main components to allow an adaptive 
behaviour.
%Mismatch-detection mechanisms can be utilised to simulate changes in the environment caused by 
%active manipulation. 
A mismatch between expected and perceived sensory effects during active manipulation 
allows to detect externally caused changes and to launch appropriate corrective actions. 
For example, as illustrated in Fig. \ref{fig:intro}, 
an initial predicted effort may not match the actual required one to lift the object
due to e.g. a change in the material or the load conditions. 
Therefore, the applied torque results in a mismatch
between the predicted and current observations.  
This mismatch is used not only to adjust future motor commands,
but also to acquire further information about the object being manipulated. 
As an example, it realizes that a filled bottle requires more efforts than an empty one
%This indicates that the system needs to adjust and correct the amount of torque in order
%to meet the expectations 
(for an early study in human control, see \cite{johansson92}). 
% MOVE FOLLOWING TO EXPERIMETAL RESULT SECTION
%Classical PID control is the most popular method to control robot joints. 
%However, it is not capable of predicting the amount of effort or torque that is required to move 
%and stay in a given position. Therefore, it is not suited for mismatch detection.
% example

% our contributions
In this work, we present a sensorimotor framework to self-learning of a specific task. 
The framework gives the ability to reach to an intended goal without any prior expert 
knowledge or assumptions about robot models.
All computations are solely based on internal motor and sensory signals that are generated 
during active interaction with the environment. 
By this, tedious calibrations can be avoided and the method can easily adapt itself to system changes,
such as a change in the task setting or of the robot itself.  

We apply the framework to learn a joint position controller of a PR2 robot. 
Given high-dimensional sensory and motor spaces and the dynamic nature of this task, 
it is a good candidate to study different aspects of the framework.
We extend our earlier work \cite{Ghadirzadeh15} by introducing advanced optimization, 
dimensionality reduction and adaptation. 
The following are the prominent features of the proposed framework:

\begin{enumerate}
\item Autonomous and incremental learning of optimal motor commands, purely based on sensorimotor 
signals in a real-time setting.
\item Automatic determination of relevant motor and sensory components to increase computational 
efficiency.
\item Active detection of changes in the environment and adaptation of the system during task
completion.
\item Additional long-term planning to optimize non-convex cost functions under some constraints.
% yeah, the additional is just for the sake of 4 A's
\end{enumerate}

\section{Related work}

A task in this work is defined as finding a sequence of actions that result in desirable 
observations or states. 
To autonomously learn such a task, Jordan and Rumelhart \cite{Jordan92} introduced distal supervised
learning. 
Similar to our approach, they attempted to overcome redundancies in the task space by using a 
trained forward model to guide a single solution. 
However, there is no guaranty for the solution to be optimal; and   
once an inverse model is learned, it is not possible to alter it to converge to a better solution. 
As we do not rely on an inverse model, but aim to find the optimal action according to a cost 
function at every time instant, we circumvent this issue.

M\"oller and Schenck \cite{moller08} combined a forward model and an inverse model to learn a 
collision-free mobile robot navigation task, as well as to distinguish dead-ends from corridors. 
Here, the forward model was used to continuously simulate sensory outcomes of a sequence of actions
such that the inverse model could be trained with the proper training data.
In contrast to our task setting, the complexity of the action space in \cite{moller08} is not high
which facilitates the performance of the inverse model. 

%Model-based reinforcement learning (RL) methods also learn a dynamic transition model to 
%maximize a long-term reward. 
Deisenroth and Rasmussen \cite{deisenroth15} introduced 
PILCO (probabilistic inference for learning control). 
PILCO applies Gaussian processes to learn forward dynamic models;
which in turn is used for policy evaluations and improvements. 
It is successfully applied to cart-pole and block stacking \cite{deisenroth11} tasks.
Instead of concentrating on policy learning, we focus here on the structure of a low-level controller 
based on sensorimotor signals. 
Our emphasis lies on the notion of adaptation 
to provide self-learning in a dynamic environment.

A method resembling our approach was used by Forss{\'e}n \cite{Forssen07}. 
Minimizing a predefined cost function predicted by the forward model, 
they learned a saccadic gaze controller.
Here, they used a kernel-based regression model to learn a visual forward model that predicts 
visual point displacements resulting from different motor commands. 
While, as in this work, no inverse model was learned, 
the aim of Forss{\'e}n was not to find optimal actions in an ill-posed problem setting,
but to speed up the training phase. 

Model predictive control (MPC) methods \cite{camacho13} follow a similar approach of using forward
dynamic models to optimize a future cost function. 
Recently, Ostafew et al.\ \cite{ostafew14} presented a mobile robot path tracking method 
based on non-linear MPC. 
Given an imperfect model of the robot, the method learns unmodelled dynamics based on Gaussian 
processes.
Similarly, Lenz et al.\ \cite{lenzdeepmpc} learned a complete set of dynamic models to master the 
task of cutting different food items based on deep learning. 
Comparable to these ideas, 
instead of optimizing the cost function over a time horizon, 
we use forward models in combination with a planning phase based on the 
RRT* algorithm to simulate action-observation pairs and determine a sequence of states towards the 
target. 
A sampling based approach as the RRT* offers good performance even in higher dimensions.

\section{Method}
\label{sec:method}
This section introduces the proposed framework for sensorimotor learning.
Fig. \ref{fig:intro} illustrates the structure of the framework.  
It consists of 
a forward model that predicts sensory outcomes of a given action and
an action planner to find a sequence of optimal actions to minimize a given cost function. 
The planner itself is divided into a global search method, that finds an optimal path to the 
goal state, and a local optimizer to steer between the states in the path. 
Furthermore, the framework addresses the problems of 
determining relevant sensory and motor channels and allows active adaptation in a dynamic
environment.

% DESCRIBE ALG1
\begin{algorithm}
\label{alg:Framework}
\SetKwInOut{Input}{input}
\caption{The framework structure to incremental model learning and reaching a goal state.}
$\mathbf{S, A, \Delta S} \gets$  MotorBabbling()\;
TrainForwardModel($\mathbf{S, A, \Delta S}$)\;
\Input{goal state $S^*$}
\While{$S_t \neq S^*$}{
	$S_t \gets$ GetCurrentState()\;
	$\mathcal{P}$ $\gets$ PlanPathToGoal($S_t$, $S^*$)\;
	\For{each state $\hat{S}^*_{t+1}$ in $\mathcal{P}$ }{		
		$A_t$ $\gets$ SteerToState($S_t, \hat{S}^*_{t+1}$)\;	  
		ExecuteAction($A_t$)\;
		$S_{t+1} \gets$ GetCurrentState()\;
		$\hat{S}_{t+1} \gets$ $\mathscr{F}(S_t, A_t) $\;
		$\zeta  \gets$ EvaluatePrediction($\hat{S}_{t+1},S_{t+1}$)\;
		\If{$\zeta < \mathscr{T}$}{
			UpdateForwardModel($S_t, A_t, \Delta S_{t}$)\;
		}   
		$S_t \gets S_{t+1}$\;
	}
 } 
\end{algorithm}
Algorithm $1$ gives an overview of the learning procedure. 
First, in an initialization phase, known as motor babbling, the robot generates some random
actions and stores them along with the resulting sensory outcomes. 
%First, in order to obtain an initial set of training samples for the learning algorithm,
%the robot generates a few random action-observation pairs, obtained through motor-babbling.
A regression model, in this case Gaussian processes (GP), is initially trained with these random samples. 
This model is then continuously updated and refined, while the robot tries to reach a given goal 
state.
%During the further procedure it is incrementally updated and optimized while reaching a given goal state. 
Here the notion of relevant dimensions and adaptation comes into play. 
The former is used to release computational overloads, while the latter is required to 
stay functional under dynamic settings. 
%Both aspects are essential for fast and precise learning. 
A long-term planner finds a path of states to the goal, that is 
compatible with some given constraints.  

%As mentioned above, our model is initially trained with data acquired in a short motor-babbling phase. 
%In this phase, random actions are generated and sensory observations of their outcome are stored.
%Importantly, we do not make any assumptions about the model of the robot or its environment. 
%The only information that is available to the system are internal motor and sensory signals. 
%Thus, our approach is initially solely based on the randomly generated training data which is 
%incrementally extended during task completion. 

%\newline

\noindent
Below follows a brief description of the functions used in Algorithm~$1$, with notations introduced in the respective sections that follow.
\newline

\noindent
\textbf{MotorBabbling()}: 
Performs a number of random actions and records the set of the action-observation pairs. 

\noindent
\textbf{TrainForwardModel}($\mathbf{S, A, \Delta S}$): 
Trains a regression model from observed outcomes $\mathbf{\Delta S}$, given applied actions $\mathbf{A}$ and current states $\mathbf{S}$ (Sec.~\ref{ssec:FMLearning}). 

\noindent
\textbf{PlanPathToGoal}($S_t, S^*$):
Finds a possible path (a number of waypoint states) from the current state $S_t$ to a goal $S^*$, that takes into account a set of constraints while planning (Sec.~\ref{ssec:Planning}). 

\noindent
\textbf{SteerToState}($S_t, S^*$):
Defining a cost function as the distance between the current state $S_t$ and goal $S^*$, 
this function finds the optimal actions, that minimize the cost (Sec.~\ref{ssec:Optimization}).

\noindent
\textbf{UpdateForwardModel}($S_t, A_t, \Delta S_t$): 
Updates the model by newly acquired data, if the prediction is poor; 
thus it incrementally improves the model (Sec.~\ref{sec:relDet} and \ref{sec:increm}).

%%%%%%%%%%%%%%%%%%%%%%%%%%%%%%%%%%%%%%%%%%%%%%%%%%%%%%%%%%%%%%%%%%%%%%%%%%%%%%%%%%%%%%%%%%%%%%%%%%%
% FORWARD MODEL LEARNING	                                        
\subsection{Forward model learning} 
\label{ssec:FMLearning}
Forward models predict the sensory outcomes of different actions of an embodied agent. 
In more general terms, they estimate how a robot's state 
will change as the result of a certain action, as the following:
\begin{equation} 
	\Delta S_t = \mathscr{F} (S_t , A_t),
	\label{eqForward}
\end{equation} 
where $S_t = [s_{t}^1, ..., s_{t}^{n_s}]$ is the state vector at time $t$, 
$\Delta S_t = S_{t+1} - S_t$ and 
$A_t=[a_{t}^1, ..., a_{t}^{n_a}]$ is the performed action at the same time step.
Here we apply GP regression to learn these sensorimotor contingencies. In our earlier works
\cite{ Ghadirzadeh15, ghadirzadeh14} we observed that GPs offer good generalization properties 
within the sensorimotor setting. 
Since, the Bayesian nature of the model takes uncertainty in the data into account, 
it is less prone to overfit and less vulnerable to noise. 
%In contrast to other methods as e.g. neural networks, 
%GPs do not require a large amount of training data nor is the slow learning of regression 
%weights needed. 

In the following we describe the steps of the forward model learning.

%%%%%%%%%%%%%%%%%%%%%%%%%%%%%%%%%%%%%%%%%%%%%%%%%%%%%%%%%%%%%%%%%%%%%%%%%%%%%%%%%%%%%%%%%%%%%%%%%%%
% GAUSSIAN PROCESS REGRESSION
\subsubsection{Gaussian process regression} 
\label{sec:GP}
For each observed state dimension $i$ a separate GP is trained, modeling a function $\mathscr{F}_i$.
Let a set of $N$ training samples be given in terms of a $N \times (n_s+n_a)$ matrix $\mathbf{X}$ with rows given by concatenated state-action pairs $X=(S,A)$, with corresponding outputs in a $N \times 1$ vector $Y^i = [ \Delta s^i ]$ . Assuming a zero-mean prior, the posterior mean of the GP is used 
as the regression output for the corresponding dimension of the test data $X_t = (S_{t}, A_{t})$,
% GP INFERENCE
%\begin{equation} % <----------------------------------- transpose is wrong?
%	\label{eq:GP_Mean}
%	\bar{y}_*^i = k(X_*,\mathbf{X})^T\mathbf{K}^{-1}Y^i		
%\end{equation}  
\begin{equation} 
	\label{eq:GP_Mean}
	\bar{y}_t^i = k(X_t,\mathbf{X})\mathbf{K}^{-1}Y^i		
\end{equation}  
and subsequently, the posterior variance gives the regression quality as
\begin{equation} 
	\label{eq:GP_Variance}
	v_t^i = k(X_t,X_t) - k(X_t,\mathbf{X}) \mathbf{K}^{-1} k(X_t,\mathbf{X})^T.
\end{equation} 
Here the vector $k(X_t, \mathbf{X})$ and matrix $\mathbf{K} = k(\mathbf{X},\mathbf{X})$ 
denote the test-train and train-train covariances respectively. 
We use the squared exponential kernel as the covariance function defined as  
\begin{equation}  
	\label{eq:covSEard}
	k(X_m, X_n) = \sigma_f \exp({\sum_{j=1}^{n_a+n_s} - \frac{\lambda_j(x_{m}^j-x_{n}^j)^2}{2} }) + \sigma_n^2 \delta_{mn},
\end{equation}  	
where $\delta_{mn}$ denotes the Kronecker delta function and
$\Theta = (\sigma_f, \sigma_n, \lambda_1, ... \lambda_{n_a+n_s})$ are hyperparameters 
which are found by optimizing the marginal log likelihood of the training data. Instead of a slowly converging gradient descent we make use of the Resilient backpropagation (Rprop) algorithm that has been shown to successfully determine hyperparameters of GPs \cite{blum13}. Rprop is a fast first order optimization method based on the sign of the local gradient and adaptive step sizes which are adjusted independently across dimensions.

To optimize the cost function the derivative of the regression output is required,
as described in the Sec. \ref{ssec:Optimization}. 
In this case, the partial derivative  of the GP posterior mean w.r.t. the $j_{th}$ dimension of the test input $X_t$ can be found as 
\begin{equation} 
	\label{eq:GP_Gradient}
	\frac{\partial \mathscr{F}_i}{\partial x_t^j}  =
	\lambda_j (X^j - x_t^j J)^T (k(X_t,\mathbf{X}) \odot (\mathbf{K}^{-1}Y^i))
\end{equation} 
%\begin{equation} % <--------------------------------------------
%	\label{eq:GP_Gradient}
%	\frac{\partial \bar{y}_*}{\partial x^*_i}  =
%	-\lambda_i^{-1} (\mathbf{X}(:,i) - x^*_i)^T (k(X_*,\mathbf{X}) \odot \mathbf{K}^{-1}Y)
%\end{equation} 
where the operator $\odot$ is an element-wise product, and $X^j$ is the $j_{th}$ column of $\mathbf{X}$, and $J$ is a $N \times 1$ vector of ones. %$\mathbf{X}(:,i)$ returns the $i_{th}$ column of the training data. 

%%%%%%%%%%%%%%%%%%%%%%%%%%%%%%%%%%%%%%%%%%%%%%%%%%%%%%%%%%%%%%%%%%%%%%%%%%%%%%%%%%%%%%%%%%%%%%%%%%%
% RELEVANCE DETERMINATION
\subsubsection{Relevance determination}
\label{sec:relDet}
Considering the limited processing resources of a robot
and a large number of degrees of freedom and sensory inputs, 
it is crucial to involve only task-relevant actions and sensory data in the computations. 
In this framework, properties of the GP regression are exploited. We use automatic relevance determination \cite{Rasmussen06} to determine which action and state dimensions are relevant for a given task. 
Considering Eq. \ref{eq:covSEard}, small values of $\lambda_j$ suggest that the $j_{th}$ input dimension is irrelevant to the predicted output $\bar{y}_t^i$ of the $i_{th}$ GP at time $t$. Therefore, this dimension can be ignored while optimizing and employing $\mathscr{F}_i$ to decrease computational efforts and reduce the level of noise. 

\begin{algorithm}
\label{alg:UpdateForwardModel}
\SetKwInOut{Input}{input}
\caption{Incremental forward model learning.}
\Input{$S_t, A_t, \Delta S_{t}$}
$X_t \gets (S_t, A_t)$,
$Y_t \gets \Delta S_{t}$ \;
\For{$i = 1,...,n_s$}{
	$r \gets$ FindRelevantDimension($\Theta_i$)\;	
%	$X^r_* \gets $ GetRelevantInputs($X_*,r$)\;
%   $X^r_* \gets $ relevant subset of $X_*$ given $r$\;
	assign $X^r_t$ as the relevant subset of $X_t$ given $r$\;
%	$\mathbf{X}^r \gets$ relevant submatrix of $\mathbf{X}$ given $r$\;
%	$\mathbf{X}^r \gets$ GetTrainingData($\mathbf{X}, r$)\;
	assign $\mathbf{X}^r$ as the relevant sub matrix of $\mathbf{X}$ given $r$\;
	$\bar{y}_t^i \gets$ GP($X^r_t, \mathbf{X}^r, Y^i, \Theta_i$)\;
	\If{$|y_t^i - \bar{y}_t^i| > \tau_q$}{
		$k_{max} \gets \max{k(X^r_t, \mathbf{X}^r)}$\;
		\If{$k_{max} > \tau_k$}{
		        $id \gets \arg\max{k(X^r_t, \mathbf{X}^r)}$\;
                  remove sample $id$ from training set of $\mathscr{F}_i$\;
%			RemoveTrainingData($id_m, i$)\;
		}
        add $(X_t, y^i_t)$ as the training sample for $\mathscr{F}_i$\;
%		AddTrainingData($X_*, y^i_*, i$)\;
	}
	$\Theta_i \gets $ TrainGP($\mathbf{X^r}, Y^i$)\;	
} 
\end{algorithm}

%\begin{algorithm}
%\label{alg:UpdateForwardModel}
%\SetKwInOut{Input}{input}
%\caption{Incremental forward model learning.}
%\Input{$S_t, A_t, \Delta S_{t}$}
%$X_* \gets [S_t, A_t]$,
%$Y_* \gets \Delta S_{t}$ \;
%\For{$y^i_* \in Y_*$}{
%	$r \gets$ FindRelevantDimension($hyp_i, i$)\;	
%	$X^r_* \gets $ GetRelevantInputs($X_*, i, r$)\;
%	$(\mathbf{X}^r, Y) \gets$ GetTrainingData($\mathbf{X}$, $i$, $r$)\;
%	$\hat{y_i} \gets$ GP($X^r_*, \mathbf{X}^r, Y, hyp_i$)\;
%	\If{$|y_i^* - \hat{y}_i| > \tau_q$}{
%		$(k_{m}, id_m) \gets \max{(k(X^r_*, \mathbf{X}))}$\;
%		\If{$k_m > \tau_k$}{
%			RemoveTrainingData($id_m, i$)\;
%		}
%		AddTrainingData($X_*, y^i_*, i$)\;
%	}
%	TrainGP($\mathbf{X^r}, Y, i$)\;	
%} 
%\end{algorithm}

%%%%%%%%%%%%%%%%%%%%%%%%%%%%%%%%%%%%%%%%%%%%%%%%%%%%%%%%%%%%%%%%%%%%%%%%%%%%%%%%%%%%%%%%%%%%%%%%%%%
% INCREMENTAL LEARNING
\subsubsection{Incremental learning and adaptation}
\label{sec:increm}
The forward model is initially trained by a few randomly generated training samples. 
In each subsequent iteration a new action-observation pair is available and a prediction is performed. A poor prediction could be
caused by either the lack of a sufficient amount of training data or by a sudden change in the environment not captured by the current model. In the former case, the model should be updated while in the latter case it needs to be adapted.
The two situations can be distinguished with the help of the test-train covariance vector $k(X_t, \mathbf{X})$, that is 
a measure of distance between the query input $X_t$ and the training data inputs $ \mathbf{X}$.
If the query input is close to a training point but the prediction is poor the model has to adapt and replace
the interfering data point with the new data sample.
Regardless of which, the current action-observation pair will be added to the training data. 

Algorithm 2 summarizes both the relevance determination and the incremental learning of the framework. A list of the functions used in algorithm can be found below, each of which is applied to the $i_{th}$ GP model, represented by $\mathscr{F}_i$.
\newline

\noindent
\textbf{FindRelevantDimension}($\Theta_i$):  
Finds the relevant dimensions $r$ of the training data, given the optimized hyperparameters $\Theta_i$ (Sec.~\ref{sec:relDet}).

%\noindent
%\textbf{GetRelevantInputs}($X_*, r$): 
%Returns the relevant subset of current input $X_*^r$.
%
%\noindent
%\textbf{GetTrainingData}($\mathbf{X}, r$): 
%Returns a matrix of the relevant training data inputs $\mathbf{X^r}$. 

\noindent
\textbf{GP}($X_t, \mathbf{X}, Y^i, \Theta_i$): 
Returns the prediction of the $i_{th}$ output dimension at a new input $X_t$,
given the training data $(\mathbf{X},Y^i)$ and hyperparameters $\Theta_i$ (Sec.~\ref{sec:GP}).

\noindent
\textbf{TrainGP}($\mathbf{X}, Y^i$): 
Optimizes $\Theta_i$ by maximizing the log likelihood of the training 
data $(\mathbf{X},Y^i)$ (Sec.~\ref{sec:GP}).

%\noindent
%\textbf{RemoveTrainingData}($id_m, i$): 
%Removes the $id_m-th$ training sample in the training data set of the  $i_{th}$ GP model, see Sec. \ref{sec:increm}.
%
%\noindent
%\textbf{AddTrainingData}($X_*, y^i_*, i$): 
%Adds the new data pair ($X_*, y^i_*$)  to the training data set of the $i_{th}$ GP model, see Sec. \ref{sec:increm}.

%%%%%%%%%%%%%%%%%%%%%%%%%%%%%%%%%%%%%%%%%%%%%%%%%%%%%%%%%%%%%%%%%%%%%%%%%%%%%%%%%%%%%%%%%%%%%%%%%%%
% Finding OPTIMUM ACTION

\begin{figure*}[t!]
	\vspace{0.3cm}
    \centering
    \includegraphics[width=1\textwidth]{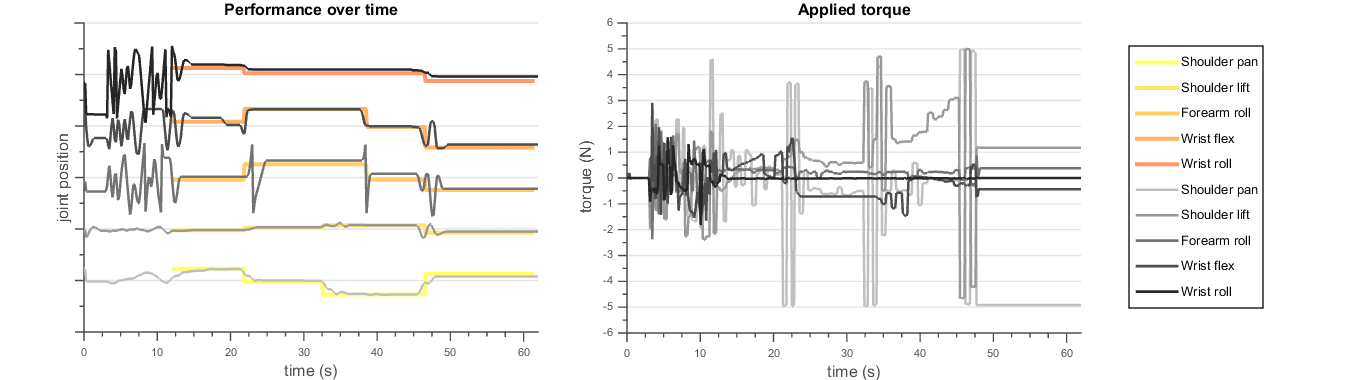}
    \caption{After an initial motor babbling period (approx. 12 s) the robot is supposed to move to different joint positions. Left: Performance over time, where the target positions are indicated by color and the actual data is shown in gray scale. For the sake of visualisation the joint positions are presented with an offset to each other.  Right: The produced efforts.}
    \label{fig:perf}
\end{figure*} 

\subsection{Optimized action selection}
\label{ssec:Optimization}

Action selection is done by finding the actions that minimize a given cost 
function defined as 
the distance between the current state $S_t$ and the goal state
$S_*$ with $\mathscr{C} = (S_t - S_*)\mathbf{W}(S_t - S_*)^T$, where $\mathbf{W}$ is a predefined diagonal weight matrix.

We assume each action dimension $a^j$ to be bounded in a symmetric range $[-\gamma_j,\gamma_j]$ as following:
\begin{equation}  
\label{eqBoundMT}
a^{j} =\gamma_j \frac{1 - \exp(-\sigma_j)}{1 + \exp(-\sigma_j)},
\end{equation}
Here $\sigma_j$ is an unbounded action parameter, which is used to minimize the cost function. 
The gradient of the cost with respect to  $\sigma_j$ is 
\begin{equation} 
	\label{eq:Gradient}
	\frac{ \partial \mathscr{C} }{\partial \sigma_j} = 
	\sum_{i=1}^{n_s} {
	\omega_i(s_*^i - s_{t}^i)
	\frac{\partial \mathscr{F}_i}{\partial a^j} }
	\frac{2\gamma_j \exp(-\sigma_j)}{(1+\exp(-\sigma_j))^2}
\end{equation}
where the $\omega_i$ are the diagonal elements of the weight matrix $\mathbf{W}$ and
$\partial \mathscr{F}_i/\partial a^j$ is given by Eq.~\ref{eq:GP_Gradient}.
The rprop algorithm is again used to minimize the cost function.

%%%%%%%%%%%%%%%%%%%%%%%%%%%%%%%%%%%%%%%%%%%%%%%%%%%%%%%%%%%%%%%%%%%%%%%%%%%%%%%%%%%%%%%%%%%%%%%%%%%
% LONG TERM PLANNING
\subsection{Long-term planning}
\label{ssec:Planning}
For long-term planning RRT* (Rapidly exploring Random Tree) \cite{Karaman11} is used as a global 
planner to find an optimal path to the goal state. 
It is well-suited to solve non-convex optimization problems under a set of constraints, 
that allow to shape the performance of the system, e.x. the overshoot or rise time.
The root of the tree is given by the current state of the robot  $S_{init} = S_t$.
At each iteration, it generates a random sample, $S_{rand}$, around the goal state.
In the case that the sampled state does not violate the given constraints, it is accepted. 
Then, the closest node in the tree, $S_{near}$, is determined.  To move from $S_{near}$ to $S_{rand}$, 
an optimized action is found by minimizing the cost function as explained in Sec. \ref{ssec:Optimization}. 
Given the optimized action and the state $S_{near}$, the forward model predicts a new state, 
$S_{new}$. 
The quality of the prediction is found according to Eq. \ref{eq:GP_Variance}.
If the quality is acceptable, $S_{new}$ will be adopted to a parent which is given by the RRT* 
method. 
An update of any parent-child relationship is performed depending on reachability and the cost of  
other points in the vicinity of $S_{new}$. 
When either the goal state can be reached from a current $S_{new}$ or 
after a size limit is reached, the optimal path through the samples space is computed. 
After determining the closest
neighbour to the goal state, its line of heritage is backtracked towards the initial state $S_{init}$. 
Finally, the forward model
is used to guide the robot along the determined path.  
\newline

In summary, we introduce a sensorimotor learning framework based on data-efficient GP regressions 
and two different optimization techniques to select goal-directed actions in a dynamic environment.
Our approach is able to handle high-dimensional motor and sensory spaces and to adapt its behaviour,
when prediction errors are too large. 
In the following, we present a number of experiments to analyse the performance of the learning 
method.

\section{EXPERIMENTS}
\label{sec:Experiments}
In this section, we exploit the introduced framework to learn a joint position controller and to
study the method in operation.
After introducing our setup (A) we provide examples of:
B) initial learning and system performance, when maturely trained, 
C) cost prediction by forward models and performance optimization, 
D) relevant dimension determination
E) adapting to load conditions and
F) long-term planning.

%%%%%%%%%%%%%%%%%%%%%%%%%%%%%%%%%%%%%%%%%%%%%%%%%%%%%%%%%%%%%%%%%%%%%%%%%%%%%%%%%%%%%%%%%%%%%%%%%%%
% SETUP	
\subsection{Experimental setup}
\label{ssec:Setup}
All experiments were performed on the right arm of a PR2 robot, including the shoulder pan (1), shoulder lift (2), forearm roll (3), wrist flex (4) and wrist roll joint (5). In the following, let $q_i$ denote the position of joint $i$,  $\dot{q}_i$ denote the velocity and the effort be denoted by $\tau_i$. Following the PR2 manual, the joint positions are limited to lie within the reachable range of radians. Furthermore, the velocity is limited to be within (-3,3) rad/s and the allowed torque is limited to  (-7,7) N for the shoulder pan and shoulder lift joints and (-3,3) N for the remaining joints to prevent damages to the robot. The entire framework is implemented in C++.  \\

Following the description in Sec. \ref{ssec:FMLearning}, the input states to the forward model at time $t$  are defined as a vector consisting of the current position, velocity and applied action or torque for each joint,  $X_t = [S_t, A_t] =  [q^1_t,... q^5_t, \dot{q}^1_t,...\dot{q}^5_t, \tau^1_t,..., \tau^5_t]$. The prediction of the forward model is the resulting state $S_{t+1}= [q^1_{t+1},... q^5_{t+1}, \dot{q}^1_{t+1},...\dot{q}^5_{t+1}]$. 

The diagonal weight matrix $\mathbf{W}$ introduced in Sec. \ref{ssec:Optimization} consists of $w_{i} = 1$ for $i = 1,...,5$ and  $w_{i} = 0.1$ for $i = 6,...,10$. By this, an error in the position of a joint has more influence than an error in the velocity.

%%%%%%%%%%%%%%%%%%%%%%%%%%%%%%%%%%%%%%%%%%%%%%%%%%%%%%%%%%%%%%%%%%%%%%%%%%%%%%%%%%%%%%%%%%%%%%%%%%%
% LEARNING CONVERGENCE	
\subsection{Learning performance}

The goal of this basic sensorimotor framework is to learn how to apply a sequence of actions to get
a desired sensory outcome. 
With this experiment, we aim to test the general learning performance of our method. 
%After a motor babbling period the robot is supposed to acquire a number of goal joint 
%configurations based on the forward model.
During a motor babbling phase, 20 randomly sampled state-action pairs are generated in order 
to initially train the forward models.
%for the method to converge quickly towards the goal state. 

As can be seen in Fig. \ref{fig:perf}, the desired configurations are reached  within only a few 
iterations after a target state is introduced. 
In spite of the large number of dimensions and the dynamic nature of the problem,
%the high level of uncertainty due to i.e. noise  
the system learns to navigate in the joint space successfully. 
At the onset of a new target, nearly all joints exhibit an overshoot behaviour. 
If such perturbations of the system are not desired a more advanced planning 
%of the optimal action sequence 
will be applied as shown in Sec. \ref{Ex:rrt}.

\begin{figure}[b!]
      \centering
      \includegraphics[width=0.39\textwidth]{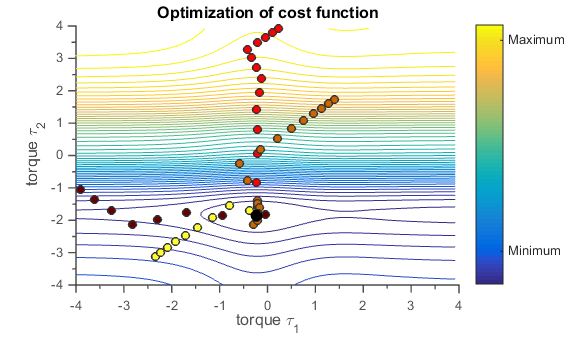}
      \caption{Four initializations for the optimization of the action with two joints. The rprop algorithm finds the minimum for several initial positions.}
      \label{fig:opt}
\end{figure} 
\begin{figure*}[t!]
	\vspace{0.3cm}
      \centering
      \includegraphics[width=0.9\textwidth]{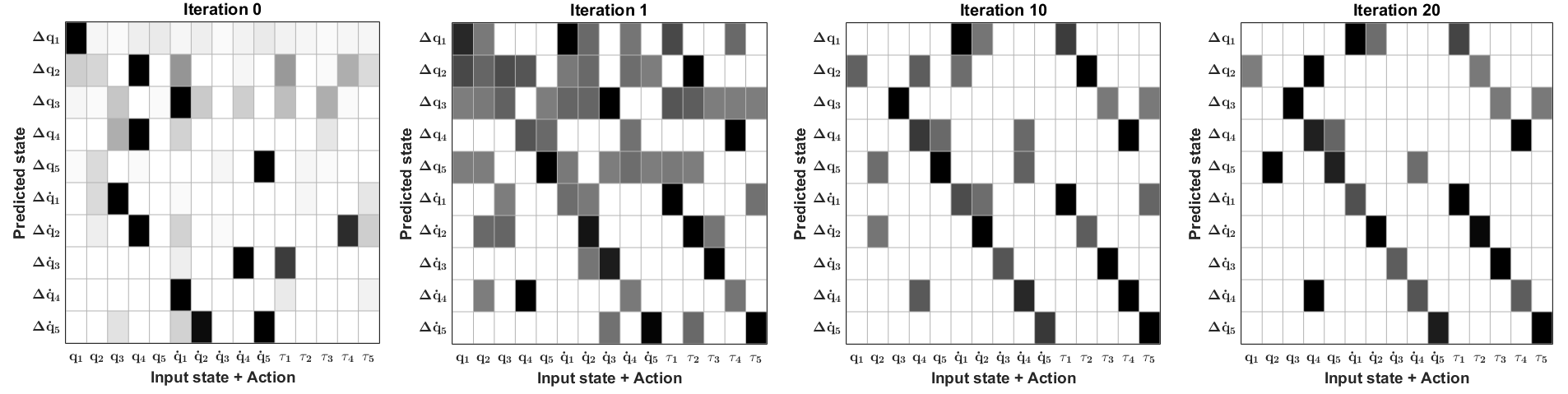}
      \caption{The normalized hyperparamters $(\lambda's)$ of each GP (row) for each input dimension (column). The darker the field, the larger the entry and the more relevant is the dimension for the GP.}
      \label{fig:hyper}
\end{figure*}

%%%%%%%%%%%%%%%%%%%%%%%%%%%%%%%%%%%%%%%%%%%%%%%%%%%%%%%%%%%%%%%%%%%%%%%%%%%%%%%%%%%%%%%%%%%%%%%%%%%
% OPTIMIZATION	
\subsection{Optimization of action selection}

Here, we investigate the nature of our gradient-based action selection approach. The goal is to acquire an understanding of the cost function and to show that, facing the pressure to react fast, the optimization can select appropriate actions. To find the optimal action while operating in up to 15 dimensions with redundant paths towards a goal state, is non-trivial. Nevertheless, the optimization algorithm applied here, rprop, is able to find minima of the cost function. While plain gradient descent moves only slowly over the shallow parts of the cost manifold, rprop is independent of the magnitude of the gradients and converges faster. 
In order to guarantee a high chance of a sufficient solution, we initialize the action parameters $\sigma_i$ at three different randomly chosen positions within the action bounds. For each initialization, we run rprop for max. 20 iterations or until convergence and apply the action with the deepest minimum. In Fig. \ref{fig:opt}, we demonstrate how several random initializations converge towards the minimum in a two-joint setting. Although the cost function might contain several local minima, especially in higher dimensions, the random initialization in each iteration is often sufficient to find appropriate actions towards a given goal state.

\begin{figure}[b!]
      \centering
      \includegraphics[width=0.49\textwidth]{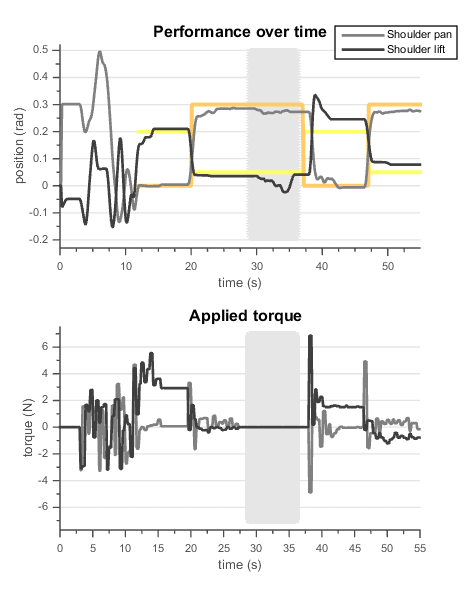}
      \caption{Adaptation to new conditions in a two-joint setting. The period under which an object is placed into the gripper is indicated by the gray area. The system learns how to adjust the torque after the change in conditions. While the same positions are aquired in both  conditions, the applied torque to reach to the target states changes before and after the loading event.}
      \label{fig:adap}
\end{figure} 

%%%%%%%%%%%%%%%%%%%%%%%%%%%%%%%%%%%%%%%%%%%%%%%%%%%%%%%%%%%%%%%%%%%%%%%%%%%%%%%%%%%%%%%%%%%%%%%%%%%
% Relevant dimensions
\subsection{Relevant dimension determination}

Our goal in this part is to explore which input dimensions are determined relevant by the forward 
models of the different states. 
Since our problem setting spans a high number of dimensions,
it is of interest to decrease this number. 
As an example, the position of the shoulder pan will not depend on the joint position of the wrist.
Including this redundant information introduces noise and does not lead to a fast and accurate 
optimization. 
Thus, the automatic relevance determination of the input dimensions is crucial to reduce 
computational loads. 
In this work, we applied the relevance determination in every iteration $t+1$, 
while only considering the set of dimensions that had been determined as relevant in the last 
iteration $t$. 
The results of this procedure are depicted in Fig. \ref{fig:hyper} for iteration 0, 1, 10 and 20 
(post motor babbling period) respectively. 
In iteration 0 it is apparent that many of the hyperparameters are very small. 
Therefore, these dimensions will not contribute significantly to the prediction. 
After removing all dimensions with hyperparameters below a small threshold, 
we observe that the system converges towards a final set of relevant dimensions. 
In most cases, $\Delta q^i_{t+1}$ and $\Delta \dot{q}^i_{t+1}$
are governed by $\dot{q}^i_{t}$ and $\tau^i_{t}$. 

\begin{figure*}[t!]
	\vspace{0.3cm}
    \centering
    \includegraphics[width=1\textwidth]{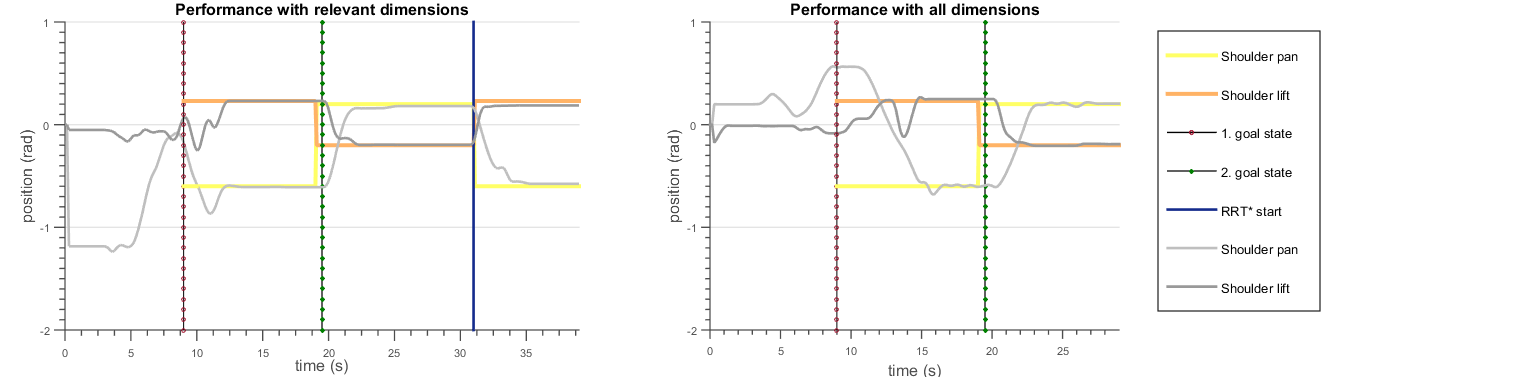}
    \caption{
	Learning performance with and without dimension reduction presented in the left and 
	right figures, respectively.  
	In the case which dimension reduction is applied, RRT* planning started after around $32$ 
	seconds (the blue line), with the aim of reducing the overshoot to zero. 	
	}
    \label{fig:perf2}
\end{figure*}

It is important to notice, that the performance depicted in Fig. \ref{fig:perf} is produced with 
dimensionality reduction.  
To better understand the importance of including only the task relevant data in the learning phase,
we compared two sample results, with and without the relevance determination.   
As shown in Fig. ~\ref{fig:perf2}, the method is applied to control only two joint positions;
as including more joints is hardly computational traceable in the latter case. 
Still, even with a two-joint setup, the difference in the performances is quite notable. 
The former case, converges faster with a more stable output. 

An interesting aspect to test in future experiments is how the relevant dimensions are 
influenced by a changing environment. 
While the joints are mostly independent in the current setting, 
an action constrained setting, such as carrying a table together with another agent, 
might change the dependences of the different joints over the time. 
In order to build an adaptive agent, these changes need to be 
detected and appropriate adjustments need to be introduced into the framework. 

%%%%%%%%%%%%%%%%%%%%%%%%%%%%%%%%%%%%%%%%%%%%%%%%%%%%%%%%%%%%%%%%%%%%%%%%%%%%%%%%%%%%%%%%%%%%%%%%%%%
% ADAPTATION
\subsection{Adaptation}

Our goal is to adaptively learn and interact in a real world setting. 
Therefore, we study here the evolution of the applied torques under changing loads. 
In order to test for this ability, we first let the arm move to two specified goal positions with the help of our framework
and  keep it stable in the later configuration. 
In a short time period an experimenter places a 
load (approx. 300 g) into the gripper of the robot without moving it considerably. 
Starting of with the previously learned forward model, the framework adapts to the new load condition 
and optimizes the applied torques in order to achieve the two previous configurations. 
Our developed method demonstrates successfully, that it is able to detect outliers in the training 
data and replace them with the pair, gathered under the new conditions. 
In Fig. \ref{fig:adap} the behaviour of the system is shown in a two-joint setting. 
Both before and after placing the load into the gripper, the desired positions are acquired. 
%More interestingly, the applied torque changes for both goal states. 
%Most attention should be paid to the applied torque.
After a new target state is set, the system tries at first to navigate using the previously learned
models. 
When this attempt is failing, as the torque is too high or low, the adaptation enables the system 
to adjust the required torque and to reach the target state after only a few iterations. 

With this experiment, we investigated only slight changes in the task setting and allowed the 
system to adjust over time. 
Since the system adapts quickly, 
it might be able to handle even more 
drastic changes within short time intervals. 
However, this remains to be tested.  

%%%%%%%%%%%%%%%%%%%%%%%%%%%%%%%%%%%%%%%%%%%%%%%%%%%%%%%%%%%%%%%%%%%%%%%%%%%%%%%%%%%%%%%%%%%%%%%%%%%
% RRT PLANNING
\subsection{Planning}
\label{Ex:rrt}

\begin{figure}[h!]
      \centering
      \includegraphics[width=0.49\textwidth]{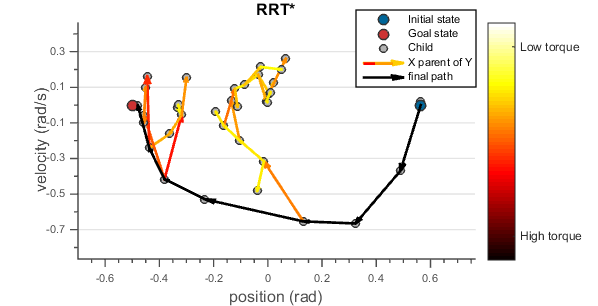}
      \caption{Path generated by the RRT* for a single joint. The color of the nodes indicates how much torque is required to reach to the specified positions. }
      \label{fig:rrt}
\end{figure}

Here we demonstrate that the RRT* algorithm successfully produces an action sequences that meets 
our specified constraint of no overshoot. 
As can be seen in Fig. \ref{fig:perf}, the basic learning framework tends to exhibit overshoots when
new references are introduced. 
%This is caused by the fact that the prediction relies on a single action step such that high 
%amounts effort are applied to reach a distant target. 
To avoid this behaviour and enable more complex movements, we introduce planning with help of the 
RRT* algorithm. 
We sample random configurations within an ellipsoid between the current state and the goal state 
with a higher probability around the goal. 
When the target can be reached from any of the samples, we follow the path
of the lowest cost. 
In Fig. \ref{fig:rrt} one example of an action tree is shown for a single joint. 
After an initial acceleration the velocity is decreased in order to come to a halt at the goal 
state. 
By altering the sampling strategy different planning behaviours can be introduced.
Fig. ~\ref{fig:perf2} (left) demonstrates an example time-domain performance to constrain 
the overshoot to zero.

\section{CONCLUSIONS AND FUTRUE WORK}

In this work, we presented a framework for sensorimotor learning based on forward models 
and two action selection methods at different hierarchy levels. 
Our approach determines relevant input dimensions at no extra cost.
%without the need of computational heavy matrix operations as required for other dimensionality reduction techniques. 
Furthermore, it is able to actively adapt to a dynamic environment and incorporate task constraints.

The experimental results provide evidence of fast, data-efficient learning in a high-dimensional action space. 
We showed that the quickly converging relevant dimensions and adaptation contribute to the efficiency and flexibility of our framework.
Long-term planning with the RRT* algorithm results in successful action generation in a constrained setting. 
%In conclusion, our approach enables the robot to learn a joint controller, that is actively adjusted if the environmental settings are altered.  

While the current framework addresses the learning of a low-level control system, in the future 
we are aiming at an integration of high-level cognitive stages. 
The presented approach can be extended to include human-robot interaction scenarios. 
Since our system is able to detect and react to external influences, 
it can learn how to interact with a human while e.x. carrying an object together.  
Mismatch detection between the prediction and outcome of sensory observations enables the agent 
to interpret the signals implicated by different forces applied by a human partner and to choose 
appropriate actions. 
This transforms the robot from a reactive and compliant partner to an active, autonomous agent.

\addtolength{\textheight}{-12cm}   % This command serves to balance the column lengths

\section*{ACKNOWLEDGMENT}

This work was supported by the EU through the project socSMCs (H2020-FETPROACT-2014) and the Swedish Research Council. 
\bibliographystyle{IEEEtran}
\bibliography{root}

\end{document}